\pdfoutput=1

\documentclass[11pt]{article}

\usepackage[review]{acl}
\usepackage{times}
\usepackage{latexsym}
\usepackage{amsmath} 
\usepackage{graphicx}
\usepackage{colortbl}
\usepackage{epsfig}
\usepackage{amssymb}
\usepackage{booktabs}
\usepackage[ruled]{algorithm2e}
\usepackage{setspace}

\usepackage{multirow}
\usepackage{color}
\usepackage{hyperref}
\usepackage{diagbox}
\usepackage[T1]{fontenc}

\usepackage[utf8]{inputenc}

\usepackage{microtype}

\usepackage{inconsolata}

\title{EBFT: Effective and Block-Wise Fine-Tuning for Sparse LLMs}
\author{
    Song Guo$^{1}$\footnotemark[2],
    Fan Wu$^{1}\footnotemark[2]$,
    Lei Zhang$^{1}$,
    Xiawu Zheng$^{1}$, \\
    \textbf{Shengchuan Zhang$^{1}$},
    \textbf{Fei Chao$^{1}$},
    \textbf{Yiyu Shi$^{2}$},
    \textbf{Rongrong Ji$^{1}$} \\
    $^1$MAC Lab, Xiamen University. $^2$University of Notre Dame.\\
    {\tt\small \{guosong, wfanstory, leizhang\}@stu.xmu.edu.cn, \{zhengxiawu, fchao, zsc\_2016, rrji\}@xmu.edu.cn}\\
    {\tt\small yshi4@nd.edu }
}

\begin{document}
\maketitle
\begin{abstract}
Existing methods for fine-tuning sparse LLMs often suffer from resource-intensive requirements and high retraining costs. Additionally, many fine-tuning methods often rely on approximations or heuristic optimization strategies, which may lead to suboptimal solutions. To address these issues, we propose an efficient and fast framework for fine-tuning sparse LLMs based on minimizing reconstruction error. Our approach involves sampling a small dataset for calibration and utilizing backpropagation to iteratively optimize block-wise reconstruction error, on a block-by-block basis, aiming for optimal solutions. Extensive experiments on various benchmarks consistently demonstrate the superiority of our method over other baselines. For instance, on the Wikitext2 dataset with LlamaV1-7B at $70\%$ sparsity, our proposed EBFT achieves a perplexity of \textbf{16.88}, surpassing the state-of-the-art DSnoT with a perplexity of 75.14. Moreover, with a structured sparsity ratio of $26\%$, EBFT achieves a perplexity of \textbf{16.27}, outperforming LoRA (perplexity 16.44). Furthermore, the fine-tuning process of EBFT for LlamaV1-7B only takes approximately \textbf{30} minutes, and the entire framework can be executed on a single \textbf{16GB} GPU. The source code is available at \href{https://github.com/sunggo/EBFT}{https://github.com/sunggo/EBFT}.
\end{abstract}

\section{Introduction}

LLMs have demonstrated remarkable potential in various NLP tasks. However, the large sizes of these models pose challenges in terms of resource requirements for deployment. For instance, the inference of GPT-3 \cite{brown2020language} in half-precision floating-point format demands at least 5 80G A100 GPUs. To address this issue, several model compression methods, such as network quantization \cite{lin2023awq,frantar2022gptq}, network pruning \cite{frantar2023sparsegpt}, and knowledge distillation \cite{hsieh2023distilling}, have been proposed to compress and accelerate these Large Language Models. Among these methods, network pruning has gained increasing attention. However, pruning often leads to a decline in the performance of sparse models. To address this issue, recent works \cite{zhang2023dynamic,frantar2023sparsegpt,zhang2023pruning} have emerged that can fine-tune the pruned models to recover their performance through regression reconstruction, costly retraining, or other heuristic methods. In this paper, we introduce EBFT, a framework designed to effectively fine-tune sparse LLMs, significantly enhancing the performance and generality of pruned models.

\textbf{Dataset used for fine-tuning}. Some existing pruning then fine-tuning approaches require significant retraining resources, partly due to the large size of the retraining dataset. For example, LLM-Pruner \cite{ma2023llm} employs Alpaca-cleaned \cite{taori2023stanford} as its fine-tuning dataset to restore the performance of sparse LLMs. Alpaca-cleaned consists of 51.8K rows of data, resulting in substantial time costs for fine-tuning LLMs. Similarly, Sheared Llama \cite{xia2023sheared} employs RedPajama \cite{together2023redpajama}, containing 2.17M rows of data, for LLM pruning and fine-tuning, which incurs huge resource costs. In this paper, we sample a small calibration dataset comprising only 256 1024-token segments extracted from C4 \cite{raffel2020exploring}. By fine-tuning sparse LLMs using these samples, we effectively reduce the resource requirements and time costs associated with the process.

\textbf{Optimization algorithm}. Current LLM pruning methods, like SparseGPT \cite{frantar2023sparsegpt}, construct reconstruction errors based on feature maps before and after pruning. They approximate the reconstruction error using the second-order term of Taylor's Formula and optimize it by regression reconstruction. Wanda \cite{sun2023simple} can be viewed as an approximation of the pruning criteria used in SparseGPT. DSnoT \cite{zhang2023dynamic} utilizes masks from Wanda or SparseGPT as initialization and designs a heuristic criterion to reselect masks that can reduce the reconstruction error further. These algorithms only optimize an approximation and often rely on heuristic experiences, leading to sub-optimal solutions. In contrast, our method defines the block-wise reconstruction error and directly optimizes it through backpropagation \cite{werbos1990backpropagation}, ensuring the attainment of an optimal and convergent solution.

\textbf{Fine-tuning costs}. EBFT can be integrated with any pruning method and optimizes the block-wise reconstruction error through a backpropagation algorithm. Our framework can avoid the simultaneous loading of all LLM blocks onto the GPU and require only a few samples, significantly reducing costs. Experimental results indicate that the time required of EBFT for fine-tuning each block in Llama-7B \cite{touvron2023llama} ranges between 50 and 60 seconds, resulting in a total time cost of approximately 30 minutes. EBFT enables fine-tuning Llama-7B with a single 16GB GPU, making LLM fine-tuning feasible even under resource-constrained conditions.

In summary, our \textbf{contributions} can be summarized as follows:
\begin{itemize}
\item We introduce EBFT, a block-by-block fine-tuning framework for sparse LLMs, which requires only a few samples, significantly reducing resource dependencies.

\item EBFT updates the network based on the minimization of block-wise reconstruction error through backpropagation, resulting in an optimal and convergent solution.

\item EBFT consistently surpass other state-of-the-art algorithms on various benchmarks and models, demonstrating the strong efficiency of our method.
\end{itemize}

\section{Related Work}
\textbf{Network pruning}. According to different levels of granularity, pruning can be categorized into unstructured pruning, structured pruning, and semi-structured pruning.
\textbf{(1) Unstructured pruning}. Unstructured pruning methods involve removing individual weights in the weight matrix. Han et al. \cite{han2015learning} proposed an algorithm based on $l_{1}$ and $l_{2}$ regulation, suggesting that smaller-norm weights are less important. LTH \cite{frankle2018lottery} increases the sparsity ratio during training and utilizes magnitude for pruning.\textbf{(2) Structured pruning}. Structured pruning involves removing entire rows or columns of the weight matrix. Li et al. \cite{li2016pruning} use the $l_{l}$-norm as the importance scores for channels. A pruning method \cite{sanh2020movement} called movement pruning was proposed, which used the product of weight value and its gradient as the criterion for importance, surpassing magnitude pruning on BERT \cite{devlin2018bert}. Cofipruning \cite{xia2022structured} generates masks for BERT pruning via l0 regularization \cite{louizos2017learning,wang2019structured}. Guo et al. \cite{guo2023automatic} analyze existing pruning criteria and propose a method based on the information bottleneck principle \cite{tishby2000information,tishby2015deep}.\textbf{(3) Semi-structured pruning}. Semi-structured pruning, also known as N:M sparsity \cite{zhou2021learning,zhang2022learning}, ensures that for every continuous M weights in the weight matrix, only N weights are non-zero. N:M sparsity can accelerate the sparse model on specific devices. Zhang et al. \cite{zhang2023bi} proposed transposable \cite{hubara2021accelerated} bi-directional masks to accelerate sparse models in both the forward and backward processes.
\begin{figure*}[thb]\label{overview}
\centerline{\includegraphics[width=0.9\linewidth]{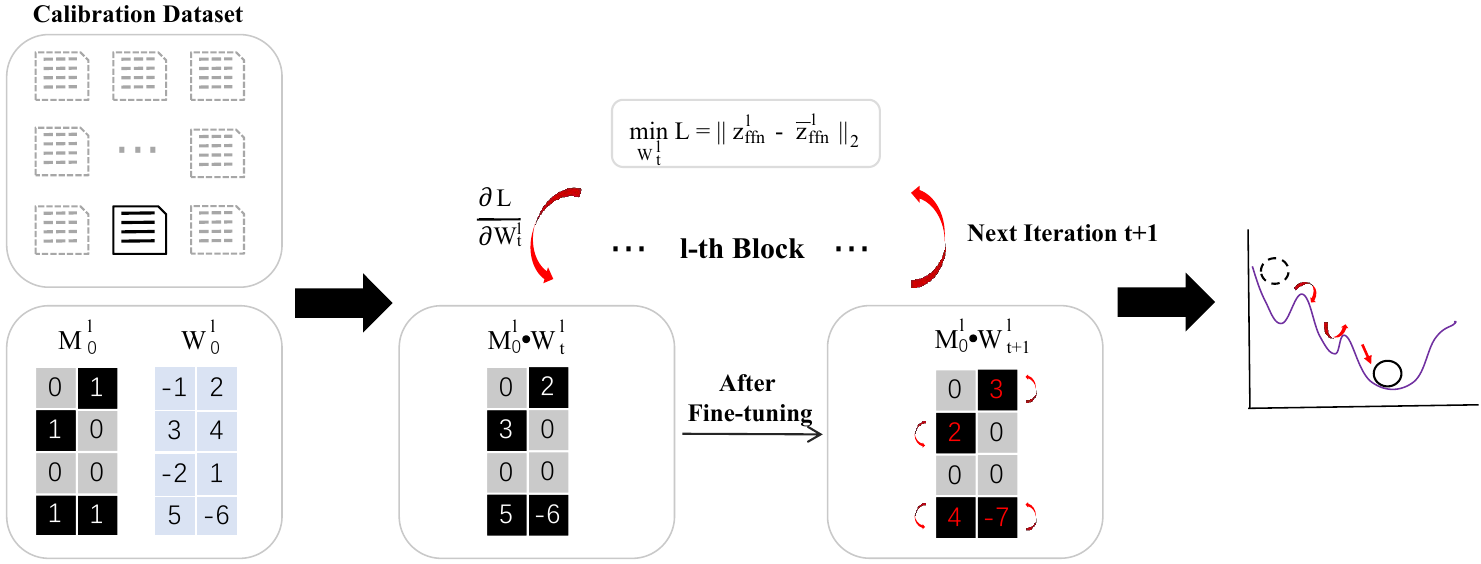}}
\caption{ EBFT can be integrated with any other pruning methods, requiring only a small number of samples from C4. When the initial mask $M_{0}^{l}$ and weight $W_{0}^{l}$ are provided, EBFT updates the weight $W_{t}^{l}$ through backpropagation to optimize the reconstruction error L mentioned in Eq.\ref{ourOO}, ultimately achieving a convergent and optimal solution. Here, $W_{t}^{l}$ represents the weight vector of the l-th block of the LLM in the t-th iteration.}
\label{fig:overview}
\end{figure*}

\textbf{Fine-tuning for pruned LLMs}. For LLMs, specific pruning methods \cite{ashkboos2024slicegpt,an2023fluctuation,syed2023prune,li2023losparse} have been proposed. LoraPruner \cite{zhang2023pruning}, LLM-pruner \cite{ma2023llm}, and Compresso \cite{guo2023compresso} aim to remove entire attention heads or FFN units in the transformers \cite{vaswani2017attention}, followed by fine-tuning on a large dataset using PEFT \cite{hu2021lora}. However, these methods suffer from performance degradation and high retraining costs. SparseGPT \cite{frantar2023sparsegpt} employs OBS \cite{hassibi1993optimal} to prune the weights of LLMs and recovers their performance through regression reconstruction. Wanda \cite{sun2023simple} proposes a new importance criterion, which approximates the criterion used in SparseGPT. DSnoT \cite{zhang2023dynamic} aims to fine-tune sparse LLMs and designs a criterion to further reduce reconstruction error by reselecting masks. These methods require costly retraining or rely on approximation and heuristic optimization strategies, resulting in significant resource consumption or sub-optimal solutions.To address these challenges, we propose a fine-tuning framework called EBFT, which helps us obtain an optimal and convergent sparse model.
\section{Methodology}
\subsection{Preliminaries}

\textbf{Large Language Model}. The structure of a large language model is based on the transformer, which consists of multiple stacked blocks. Each block consists of two modules: multi-head self-attention (MHA) and multi-layer perceptron (MLP). MHA typically comprises four linear layers, while MLP consists of two or three linear layers. For the l-th block in the large language model, it can be formulated as follows:
\begin{equation}
\begin{aligned}
    z_{attn}^{l} = MHA(W_{mha}^{l},LN(z_{ffn}^{l-1}))+z_{ffn}^{l-1},\\
    z_{ffn}^{l} = MLP(W_{mlp}^{l},LN(z_{attn}^{l}))+z_{attn}^{l},
\end{aligned}
\end{equation}
where $W_{mha}^{l}$ represents the weight vector of the multi-head self-attention module, and $W_{mlp}^{l}$ represents the weight vector of the multi-layer perceptron module in the i-th block. LN represents the layer normalization function. $z_{ffn}^{l-1}$ denotes the output of the (l-1)-th block, which serves as the input to the l-th block. The input $z_{ffn}^{l-1}$ is first passed through the MHA module and then through the MLP module.

\textbf{Pruning for LLMs.} Existing pruning methods for LLMs \cite{frantar2023sparsegpt,zhang2023dynamic,bovza2024fast,das2023beyond} typically employ the reconstruction error of the layer-wise feature maps before and after pruning as the optimization objective. This objective can be defined as follows:
\begin{equation}\label{reconstruction}
    \min_{M,\bar{W}} ||WX-(M \odot\bar{W} )X||_2,\;s.t.\;1-\frac{||M||_0}{N} = S,
\end{equation}
where X represents the input activation. W and $\bar{W}$ represent the original and remaining weight vectors, respectively, of any layer in the block of the LLM. M$\in\left\{0,1\right\}^{N}$ is the mask for this layer, indicating whether the corresponding weights should be preserved (1) or discarded (0). S is the pre-designed target sparsity, and N denotes the total number of weights in the layer.

These methods often employ the second-order term of the Taylor formula to approximate the layer-wise reconstruction error in Eq.~\ref{reconstruction} or design heuristic criteria to optimize Eq.~\ref{reconstruction}. However, these approaches may result in suboptimal solutions.
\subsection{EBFT}
\textbf{Overview}. We propose a framework called EBFT for the fine-tuning of sparse LLMs, aiming to achieve optimal solutions. Unlike other costly methods that involve pruning and then fine-tuning on a large dataset \cite{xia2023sheared,ma2023llm,zhang2023pruning}, EBFT only requires a small calibration dataset consisting of a few samples. Specifically, we extract 256 1024-token samples from C4 and use them as the calibration dataset denoted as $D_c$. The principle of EBFT is based on minimizing the block-wise reconstruction error. An overview of our algorithm is depicted in Fig.~\ref{fig:overview}.

\textbf{Optimization objective}. For the l-th block in the sparse LLM, it can be formulated as:
\begin{equation}
    \begin{aligned}
        \bar{z}_{attn}^{l} = MHA(\bar{W}_{mha}^{l},LN(\bar{z}_{ffn}^{l-1}))+\bar{z}_{ffn}^{l-1},\\
        \bar{z}_{ffn}^{l} = MLP(\bar{W}_{mlp}^{l},LN(\bar{z}_{attn}^{l}))+\bar{z}_{attn}^{l},
    \end{aligned}
\end{equation}
where $\bar{W}_{mha}^{l}=M_{mha}^{l}*W_{mha}^{l}$ and $\bar{W}_{mlp}^{l}=M_{mlp}^{l}*W_{mlp}^{l}$ represent the remain weight vector of the multi-head self-attention module and multi-layer perceptron module, respectively, in the l-th block. $M_{mha}^{l}$ and $M_{mlp}^{l}$ represent their corresponding masks. $\bar{z}_{ffn}^{l}$ denotes the output of the l-th block after pruning. 

we define our block-wise optimization objective as:
\begin{equation}\label{ourOO}
\min_{\bar{W}_{mha}^{l},\bar{W}_{mlp}^{l}}|| z_{ffn}^{l}-\bar{z}_{ffn}^{l}||_2
\end{equation}


In Eq.~\ref{ourOO}, we preserve the masks obtained from other pruning methods unchanged and focus on optimizing the remaining weights within the current block.

Compared to the layer-wise reconstruction error in Eq.\ref{reconstruction}, the block-wise optimization process in Eq.\ref{ourOO} allows for interaction and information exchange among different layers within the block. This enables the model to avoid potential issues associated with local optima in layer-wise optimization and explore the solution space more effectively, leading to the discovery of a globally optimal solution. Our EBFT is to directly optimize Eq.\ref{ourOO} without relying on any approximations or heuristic methods.

\textbf{Optimization algorithm}. Unlike some methods \cite{kwon2022fast,frantar2023sparsegpt,zhang2023pruning} that update the weights of LLM based on regression reconstruction or costly retraining, we employ the backpropagation algorithm to minimize Eq.~\ref{ourOO} by updating the value of the variable $\bar{W}_{mha}^{l}$ and $\bar{W}_{mlp}^{l}$ block by block on the $D_c$, without utilizing any heuristic methods. 

The workflow of our EBFT framework is illustrated in Alg.~\ref{alg:EBFT}. Prior to the fine-tuning process, we establish a maximum iteration T to control the overall fine-tuning cost. Specifically, we set T to 10 epochs. During the fine-tuning phase, if the loss remains unchanged or changes within a small range, we consider the loss to have converged. At this point, the fine-tuning algorithm for the current block will terminate early, allowing us to proceed to the subsequent block for a new round of fine-tuning.    

In Alg.~\ref{alg:EBFT}, $m_{0}$ can be obtained from any pruning methods. $\alpha$ represents the learning rate which determines the size of updating step for the variable $\bar{W}_{l}^{t}$. Specifically, we set the learning rate to 2e-4.

\begin{algorithm}[t]
\small
\SetKw{return}{return}
\SetKwData{Left}{left}\SetKwData{In}{in}
\SetKwInOut{Input}{input}
\SetKwInOut{Output}{output}

	\Input{sparse LLM F with L blocks; Initial Mask $m_{0}$; Calibration dataset $D_{c}$; Max fine-tuning iterations T; Learning rate $\alpha$;
 } 
	\Output{Fine-tuned sparse LLM $F_{T}$}
        \For{block $l=1$ \KwTo $L$}{
        \For{iteration $t=0$ \KwTo $T$}{
            $E\leftarrow$ Calculating the reconstruction error via Eq.~\ref{ourOO}.\\
            \textbf{If} E is \textbf{convergent:}\\
            \hspace{2em}\textbf{break}\\
            $\nabla \bar{W_{t}^{l}}\leftarrow$ Calculating the gradient of $\bar{W_{t}^{l}}$ with respect to E through Bp algorithm. \\
            $\bar{W}_{t+1}^{l}\leftarrow$ $\bar{W_{t}^{l}}$ -$\alpha$$\nabla \bar{W_{t}^{l}}$ \\ 
            }
            }
        \return$F_{T}$;
 	 \caption{Pseudocode of EBFT} 
   \label{alg:EBFT}
 	 \end{algorithm}

\section{Experiments}
\textbf{Models and Baselines.} We apply magnitude pruning, SparseGPT, and Wanda techniques to the widely adopted LLMs, LlamaV1 \cite{touvron2023llama} and LlamaV2 \cite{touvron2023llama2}. Subsequently, we compare the evaluation results of the state-of-the-art method DsnoT \cite{zhang2023dynamic} with our approach on the pruned LlamaV1 and LlamaV2, considering both unstructured sparsity and N:M sparsity. To further assess the effectiveness of our method, we also compare EBFT with LoRA \cite{hu2021lora} under structured sparsity using FLAP \cite{an2023fluctuation}.

\textbf{Evaluation.} To evaluate the performance of our method and other baselines, we conduct comparisons on the widely-used dataset Wikitext2 \cite{merity2016pointer} to calculate perplexity scores. Additionally, we perform a series of zero-shot tasks, including PIQA \cite{bisk2020piqa}, StoryCloze \cite{mostafazadeh2017lsdsem}, ARC-Easy and ARC-Challenge \cite{clark2018think}, HellaSwag \cite{zellers2019hellaswag}, Winogrande \cite{sakaguchi2021winogrande}, and Boolq \cite{clark2019boolq}. These tasks aim to assess the generality of the pruned model.
\begin{table*}[t]
\footnotesize
\begin{center}
  \begin{tabular}{cccccc|cccccc}
    \toprule
    \space &\multicolumn{5}{c|}{\textbf{LlamaV1-7B}} & \multicolumn{5}{c}{\textbf{LlamaV2-7B}}\\
    \midrule
    \multicolumn{1}{c|}{\diagbox[width=7em]{Method}{Sparsity}}&\multicolumn{1}{c}{$\boldsymbol{50\%}$}&\multicolumn{1}{c}{$\boldsymbol{60\%}$}&\multicolumn{1}{c}{$\boldsymbol{70\%}$}&\multicolumn{1}{c}{$\boldsymbol{80\%}$}&\multicolumn{1}{c|}{$\boldsymbol{90\%}$}&\multicolumn{1}{c}{$\boldsymbol{50\%}$}&\multicolumn{1}{c}{$\boldsymbol{60\%}$}&\multicolumn{1}{c}{$\boldsymbol{70\%}$}&\multicolumn{1}{c}{$\boldsymbol{80\%}$}&\multicolumn{1}{c}{$\boldsymbol{90\%}$}\\
    \midrule
    \multicolumn{1}{c|}{Magnitude} & 17.29 & 559.99 & 48415 & 132176 & 317879 & 16.03 & 1924.81 & 49906 & nan & nan\\
    \multicolumn{1}{c|}{w. DsnoT} & 13.80 & 127.67 & 9614795 & 37474 & 202562 & 13.90 & 3749.55 & 14271e4 & 21760e2 & 34462e2\\
    \rowcolor{gray!=25}\multicolumn{1}{c|}{\textbf{w. Ours}} & \textbf{7.11} & \textbf{9.53} & \textbf{26.30} & \textbf{659.12} & \textbf{9718.99} & \textbf{6.59} & \textbf{9.29} & \textbf{33.50} & \textbf{462.32} & \textbf{2930.51}\\
    \midrule
    \multicolumn{1}{c|}{Wanda} & 7.26 & 10.69 & 88.84 & 5671.52 & 12748 & 6.94 & 10.96 & 78.26 & 3136.23 & 6995.88\\
    \multicolumn{1}{c|}{w. DsnoT} & 7.14 & 10.40 & 75.14 & 3635.94 & 9043.63 & 6.85 & 10.85 & 75.55 & 4197.74 & 7311.58\\
    \rowcolor{gray!=25}\multicolumn{1}{c|}{\textbf{w. Ours}} & \textbf{6.81} & \textbf{8.59} & \textbf{16.88} & \textbf{118.38} & \textbf{2993.32} & \textbf{6.18} & \textbf{7.90} & \textbf{16.94} & \textbf{72.80} & \textbf{903.45} \\
    \midrule
    \multicolumn{1}{c|}{SparseGPT} & 7.20 & 10.40 & 27.00 & 167.55 &  3912.78& 7.09 &10.54 & 29.37&131.17&1542.22\\
    \multicolumn{1}{c|}{w. DsnoT} & 9.25 & 9.68 & 46.99 & 8038.14 & 198898 & 6.97 & 10.23 & 59.62 & 2510.54 & 49639 \\
    \rowcolor{gray!=25}\multicolumn{1}{c|}{\textbf{w. Ours}} & \textbf{6.73} & \textbf{8.33} & \textbf{16.07} & \textbf{141.15} & \textbf{3366.39} & \textbf{6.20} & \textbf{7.88} & \textbf{18.13} & \textbf{130.89} & \textbf{1233.80}\\
    \bottomrule
  \end{tabular}
\end{center}
\caption{Comparison of perplexity for pruning and fine-tuning LlamaV1-7B and LlamaV2-7B on Wikitext2 dataset at unstructured sparsity levels ranging from $50\%$ to $90\%$.}
\label{tab:un_wiki}
\end{table*}
\subsection{Language Modeling}
\textbf{Unstructured Pruning.} We perform comprehensive comparative experiments on the Wikitext2 dataset, and the results are presented in Table.\ref{tab:un_wiki} We compare the perplexity of pruned LlamaV1 and LlamaV2 models using our method, DsnoT, magnitude pruning, Wanda, and SparseGPT across a range of sparsity levels, from $50\%$ to $90\%$. The experimental results show the strong effectiveness of our EBFT. We can observe that regardless of the magnitude pruning method used, be it SparseGPT or Wanda, our method enhances the performance of the sparse model. For instance, with magnitude pruning, our method achieves a perplexity of 7.11, surpassing the perplexity of 17.29 before fine-tuning, and even outperforming Wanda (7.26) and SparseGPT (7.20).

We also find that as the sparsity increases, two observations emerge:
(1) The state-of-the-art DsnoT loses its effectiveness as a fine-tuning method. For example, when using SparseGPT, DsnoT degrades the performance of the sparse model at sparsity levels of $70\%$, $80\%$, and $90\%$. This demonstrates the limitations of heuristic optimization strategies, which lack theoretical support.
(2) The advantage of our method becomes more pronounced, indicating that our method enhances the ability of pruned models even at extremely high sparsity levels.

In Table \ref{tab:un_wiki}, we further observe that SparseGPT, which updates the values of the remaining weights, outperforms Wanda, which leaves the remaining weights unchanged. As sparsity increases, the advantage of SparseGPT over Wanda becomes more evident, particularly at high sparsity levels. Additionally, the DsnoT approach, which reselects the masks after pruning and keep weights unchanged, also faces challenges. For example, when the sparsity exceeds $70\%$, regardless of LlamaV1 or LlamaV2, DsnoT significantly decreases the performance of the sparse model pruned by SparseGPT. In contrast, our method effectively and efficiently fine-tunes the weights of the LLM block by block, surpassing other baselines overall. In the later section, we will conduct comprehensive and detailed experiments to further compare mask-tuning and weight-tuning.
\begin{table}[t]
\footnotesize
\centering
\begin{tabular}{ccc|cc}
\toprule
\multicolumn{1}{c}{} & \multicolumn{2}{c|}{\textbf{LlamaV1-7B}} & \multicolumn{2}{c}{\textbf{LlamaV2-7B}} \\
\midrule
\multicolumn{1}{c}{\diagbox[width=6em]{Method}{Sparsity}} & $\boldsymbol{2:4}$ & $\boldsymbol{4:8}$ & $\boldsymbol{2:4}$ & $\boldsymbol{4:8}$ \\
\midrule
 \multicolumn{1}{c|}{Magnitude}& 42.54 & 16.83&54.39 &16.53 \\
 \multicolumn{1}{c|}{w. DsnoT}& 38.32& 17.01&40.81 &18.34  \\
 \rowcolor{gray!=25}\multicolumn{1}{c|}{\textbf{w. Ours}} &\textbf{9.62} &\textbf{8.10} & \textbf{9.14}& \textbf{7.56} \\
 \midrule
 \multicolumn{1}{c|}{Wanda}&11.50 &8.57 &12.11 &8.66 \\
 \multicolumn{1}{c|}{w. DsnoT}&10.95 &8.46 &11.98 &8.57  \\
 \rowcolor{gray!=25}\multicolumn{1}{c|}{\textbf{w. Ours}} &\textbf{8.89} &\textbf{7.66} &\textbf{8.30} &\textbf{7.11} \\
 \midrule
 \multicolumn{1}{c|}{SparseGPT}&11.05 &8.55 & 10.44&8.01\\
 \multicolumn{1}{c|}{w. DsnoT}&10.00 &8.26 &10.06 &8.06\\
 \rowcolor{gray!=25}\multicolumn{1}{c|}{\textbf{w. Ours}} &\textbf{8.82} &\textbf{7.59} &\textbf{8.25} &\textbf{7.06}\\
 
\bottomrule
\end{tabular}
\caption{Comparison of perplexity for pruning and fine-tuning LlamaV1-7B and LlamaV2-7B on the Wikitext2 dataset at N:M sparsity levels, including two patterns, 2:4 and 4:8.}
\label{tab:nm_wiki}
\end{table}

\textbf{Semi-structured Pruning.} Semi-structured pruning, also known as N:M sparsity, is considered superior to unstructured pruning when it comes to accelerating models on devices. We conducted extensive comparison experiments on the Wikitext2 dataset, and the results are presented in Table \ref{tab:nm_wiki}. Irrespective of the 2:4 or 4:8 pattern, our method consistently outperforms DsnoT, significantly enhancing the performance of the pruned models. For example, when using the 2:4 pattern and Wanda mask initialization, our method achieves a perplexity of 8.30 for the sparse LlamaV2 model, which even surpasses the performance of DsnoT using the 4:8 pattern. The sparse LLMs pruned by magnitude pruning and fine-tuned with our method demonstrate a remarkable improvement. Our fine-tuning approach can effectively narrow the performance gap between magnitude pruning and the state-of-the-art baselines, Wanda and SparseGPT. 
\begin{table*}[t]
\footnotesize
\begin{center}
  \begin{tabular}{cccccccccc}
    \toprule
    \textbf{Model}&\textbf{Method}&\textbf{PIQA}&\textbf{ARC-E}&\textbf{ARC-C}&\textbf{WinoGrande}&\textbf{HellaSwag}&\textbf{Boolq}&\textbf{StoryCloze}&\textbf{Mean}\\
    \midrule
    \multirow{8}{*}{\textbf{Lla.1(60\%)}}&Mag.&60.55&42.30&23.21&50.04&31.86&38.29&57.40&43.38\\
    &w.DSnoT&66.65 &51.01 &26.02 &52.96 &38.31 &46.82 &65.37 &49.59\\
    \rowcolor{gray!=10}&\textbf{w.Ours}&\textbf{72.69} &\textbf{63.26} & \textbf{32.17}&\textbf{63.85} &\textbf{46.61} & \textbf{65.72}& \textbf{73.33}&\textbf{59.66}\\
    &Wanda&72.74 &62.67 &30.03 &62.67 &43.71 & 68.90&71.25 &58.85\\
    &w.DSnoT&73.07 &63.38 &30.80 &61.56 &43.51 & 68.20&71.46 &58.85\\
    \rowcolor{gray!=10}&\textbf{w.Ours}&\textbf{73.67} &\textbf{65.57} &\textbf{32.17} &\textbf{65.11} &\textbf{47.80} & \textbf{69.79}&\textbf{73.86} &\textbf{61.14}\\
    &SparseGPT&72.36 &62.58 &31.14&64.40 &45.38 & \textbf{69.79}&73.65 &59.90\\
    &w.DSnoT&73.70 &63.17 &31.83&63.06 &47.41 &67.52&73.22&59.99\\
    \rowcolor{gray!=10}&\textbf{w.Ours}&\textbf{73.77} &\textbf{64.02} &\textbf{32.51}&\textbf{64.40} &\textbf{47.84} &69.27&\textbf{74.02}&\textbf{60.83}\\
    \midrule
    \multirow{8}{*}{\textbf{Lla.2(60\%)}}&Mag.&62.73&44.78&25.00&53.12&34.99&47.86&62.21&47.24\\
    &w.DSnoT&69.42 &63.13&30.89&61.56&40.48&54.77&67.93&55.46\\
    \rowcolor{gray!=10}&\textbf{w.Ours}&\textbf{72.63} &\textbf{64.94} & \textbf{32.25}&\textbf{65.11} &\textbf{46.40} & \textbf{71.01}& \textbf{73.06}&\textbf{60.77}\\
    &Wanda&71.71 &64.98 &30.55&\textbf{64.56} &43.82 & 65.57&71.99 &59.02\\
    &w.DSnoT&71.33 &64.44 &29.95&64.17 &42.53 & 64.83&70.55 &58.25\\
    \rowcolor{gray!=10}&\textbf{w.Ours}&\textbf{73.56}&\textbf{68.73} &\textbf{33.19}&64.40 &\textbf{47.26} & \textbf{67.22}&\textbf{73.49} &\textbf{61.12}\\
    &SparseGPT&71.44 &63.72 &31.48&66.69 &45.25 & 72.54&\textbf{74.29} &60.77\\
    &w.DSnoT&72.85 &66.58 &33.19&62.83 &46.71 &65.72&73.54&60.20\\
    \rowcolor{gray!=10}&\textbf{w.Ours}&\textbf{73.29} &\textbf{67.42} &\textbf{32.59}&\textbf{66.98} &\textbf{47.10} &\textbf{72.60}&73.70&\textbf{61.96}\\
    \midrule
    \multirow{8}{*}{\textbf{Lla.1(2: 4)}}&Mag.&68.01&53.32&27.22&59.91&42.30&53.09&70.02&53.41\\
    &w.DSnoT&68.18 &54.38 &26.54 &58.96 &41.24 &48.32 &68.68 &52.33\\
    \rowcolor{gray!=10}&\textbf{w.Ours}&\textbf{72.80} &\textbf{64.18} & \textbf{30.89}&\textbf{64.25} &\textbf{45.80} & \textbf{68.29}& \textbf{72.37}&\textbf{59.80}\\
    &Wanda&70.40 &60.82 &27.79 &63.22 &42.08 & 69.08&70.71 &57.76\\
    &w.DSnoT&70.62 &61.78 &28.07 &61.56 &42.35 & 48.32&70.71 &54.91\\
    \rowcolor{gray!=10}&\textbf{w.Ours}&\textbf{72.42} &\textbf{64.81} &\textbf{30.97} &\textbf{65.19} &\textbf{46.05} & \textbf{67.25}&\textbf{72.42} &\textbf{59.87}\\
    &SparseGPT&71.22&60.73 &30.46&63.38 &42.95 & \textbf{69.85}&70.23 &58.40\\
    &w.DSnoT&72.63 &63.13 &30.72&62.67 &45.91 &67.77&71.73&59.22\\
    \rowcolor{gray!=10}&\textbf{w.Ours}&\textbf{73.45} &\textbf{64.77} &\textbf{30.80}&\textbf{66.30} &\textbf{46.39} &68.44&\textbf{72.15}&\textbf{60.33}\\
    \midrule
    \multirow{8}{*}{\textbf{Lla.2(2: 4)}}&Mag.&70.08&61.91&30.12&60.93&\textbf{45.43}&59.85&72.31&57.23\\
    &w.DSnoT&69.10 &61.45 &29.01 &59.12&43.75 &65.37&70.82&55.76\\
    \rowcolor{gray!=10}&\textbf{w.Ours}&\textbf{73.07} &\textbf{67.17} & \textbf{30.72}&\textbf{64.64} &45.27 & \textbf{66.73}& \textbf{72.63}&\textbf{60.03}\\
    &Wanda&70.89 &61.91 &30.72&62.51&41.27 & 68.53&70.23&58.01\\
    &w.DSnoT&70.18&61.74&29.78&62.75&40.90&67.86&69.91 &57.59\\
    \rowcolor{gray!=10}&\textbf{w.Ours}&\textbf{72.91}&\textbf{65.91} &\textbf{31.91}&\textbf{63.77} &\textbf{45.49} & \textbf{69.33}&\textbf{73.06} &\textbf{60.34}\\
    &SparseGPT&70.40&63.80&31.23&65.75&43.83&68.04&73.06 &59.44\\
    &w.DSnoT&73.34 &65.24 &\textbf{32.17}&63.14&45.41&67.55&73.76&60.09\\
    \rowcolor{gray!=10}&\textbf{w.Ours}&\textbf{73.34} &\textbf{66.33} &30.80&\textbf{65.88} &\textbf{45.80} &\textbf{69.79}&\textbf{73.76}&\textbf{60.82}\\
    
    \bottomrule
  \end{tabular}
\end{center}
\caption{Accuracy results of pruning and fine-tuning LlamaV1-7B and LlamaV2-7B on a series of zero-shot tasks at $\textbf{60\%}$ sparsity and \textbf{2:4} pattern sparsity.}
\label{tab:zeroshot}
\end{table*}

\subsection{Zero-Shot Tasks}
We conducted extensive experiments to evaluate the performance of the sparse model on 7 zero-shot tasks. The metric we used is accuracy. The experimental results of different methods at the unstructured sparsity level are shown in Table \ref{tab:zeroshot}. It can be observed that EBFT significantly enhances the generality of the pruned model. For instance, with magnitude pruning, EBFT improves the accuracy by 16.28 on LlamaV1-7B and by 13.53 on LlamaV2-7B. With Wanda, our methods achieve a mean accuracy of 61.14 on LlamaV1-7B and 61.12 on LlamaV2-7B. However, DSnoT hardly enhances the performance of the pruned model. It achieves a mean accuracy of 58.85 on LlamaV1-7B and 58.25 on LlamaV2-7B, respectively. For LlamaV2, DSnoT even degrades the performance after fine-tuning. The mean accuracy before fine-tuning for Wanda and SparseGPT is 59.02 and 60.77, respectively. After fine-tuning, the mean accuracy drops to 58.25 for Wanda and 60.20 for SparseGPT, highlighting the limitations of DSnoT. In contrast, after fine-tuning with EBFT, the sparse LlamaV2 model shows a significant improvement in overall accuracy, with a mean accuracy of 61.12 for Wanda and 61.96 for SparseGPT.

\textbf{N:M sparsity.} We also investigated the generality of our EBFT approach at N:M sparsity levels. Similar to unstructured sparsity, EBFT demonstrates significant advantages compared to other baselines. The experimental results for the 2:4 pattern are presented in Tab.\ref{tab:zeroshot}. In the case of magnitude pruning, EBFT improves the mean accuracy of sparse LlamaV1-7B by 6.39 and sparse LlamaV2-7B by 2.8. Conversely, DSnoT fails to restore the performance of magnitude-pruned sparse models. When using Wanda initialization, EBFT enhances the mean accuracy of sparse LlamaV1-7B by 2.11 and sparse LlamaV2-7B by 2.33. Under SparseGPT initialization, EBFT improves the mean accuracy of sparse LlamaV1-7B by 1.93 and sparse LlamaV2-7B by 1.38. In contrast, DSnoT loses its effectiveness with the current pattern as a fine-tuning method. Excluding the SparseGPT initialization on LlamaV2-7B, DSnoT significantly degrades the accuracy of the sparse model. For instance, with Wanda initialization, it results in a drop of 2.85 in accuracy for LlamaV1-7B and 0.42 for LlamaV2-7B.  
\begin{figure}[thb]\label{nsamples}
\centerline{\includegraphics[width=0.9\linewidth]{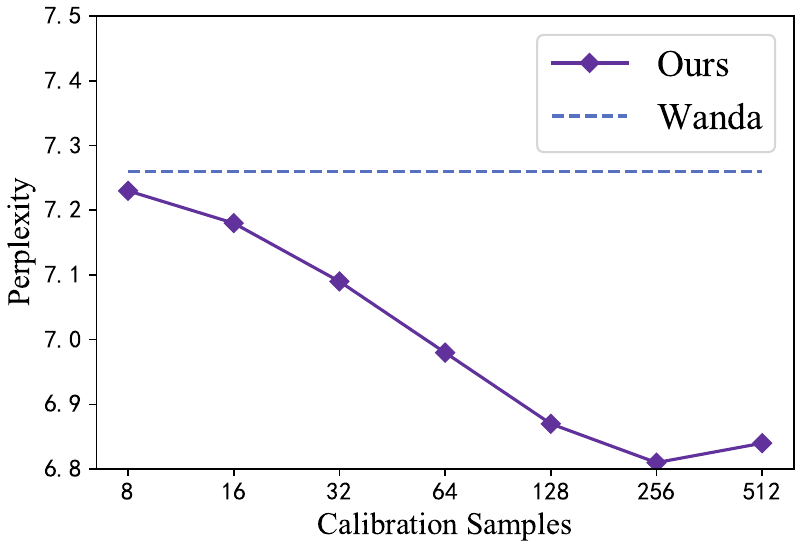}}
\caption{The perplexity of the fine-tuned LlamaV1-7B on Wikitext2, with a sparsity level of 50\%, varies with the number of samples in the calibration dataset.}
\label{fig:nsamples}
\end{figure}
\subsection{Calibration Samples}
We vary the number of samples in the calibration dataset and generate a plot illustrating the perplexity and number of samples for the fine-tuned sparse LlamaV1-7B under Wanda initialization. The results are presented in Fig.\ref{fig:nsamples}. The experimental findings demonstrate the robustness of our proposed method. Generally, as the number of samples increases, the performance of the sparse model improves. However, once the number of samples reaches 512, the perplexity does not decrease further. Notably, even with just 8 samples, the fine-tuned sparse LlamaV1 model exhibits an improvement compared to the model before fine-tuning. In addition, as the number of samples decreases, the fine-tuning speed can be further accelerated. 
\subsection{EBFT vs. LoRA}
Low-Rank Adaptation (LoRA) has gained popularity as a technique for retraining large language models. Recent works such as \cite{ma2023llm,guo2023compresso,li2023losparse} have extensively used LoRA for retraining pruned models. This involves fine-tuning the low-rank parameters A and B of an additional adapter on a large dataset to restore its performance. In this paper, we provide a detailed comparison of the fine-tuning cost and performance between LoRA and our EBFT.

In our study, we applied Low-Rank Adaptation (LoRA) and EBFT to FLAP \cite{an2023fluctuation} with structured sparsity levels. FLAP is a state-of-the-art method that outperforms LLM-Pruner in various tasks. It introduces a novel metric for channels in large language models and utilizes this metric score to search for the global structure of the model. We utilized the masks generated by FLAP as initialization for the fine-tuning process.

When fine-tuning the model pruned by FLAP using LoRA, we selected the Alpaca-GPT4 dataset as the retraining dataset. The Alpaca-GPT4 dataset consists of 50k rows of data and was fine-tuned using GPT4. We performed fine-tuning with LoRA for 2 epochs on the Alpaca-GPT4 dataset, using a learning rate of 1e-4 and a batch size of 64, which is the same as LLM-Pruner.

The fine-tuning methods employed in recent state-of-the-art works, as mentioned above, can incur a significant retraining cost. We compared their retraining methods with ours on a 40G A100 GPU. The time costs and perplexity on Wikitext2 of LoRA and EBFT are listed in Table 4. It is observed that compared to LoRA, our EBFT achieves a 10$\times$ speedup, resulting in a significant reduction in fine-tuning costs. Additionally, EBFT demonstrates better performance compared to LoRA. As shown in Table \ref{tab:time-cost}, when reducing 20\% of the parameters of LlamaV2-7B, EBFT achieves a perplexity of 15.71 on Wikitext2, which is superior to the perplexity obtained by LoRA (16.08).
\begin{table}[h]
\centering
\begin{center}
\begin{tabular}{cccc}
\hline
\textbf{Method} & \textbf{sparsity}&\textbf{time}& \textbf{perplexity} \\
\hline
LoRA & 20\% & 5h & 16.08 \\
\hline
\textbf{Ours} & 20\% & \textbf{0.5h} & \textbf{15.71} \\
\hline
\end{tabular}
\end{center}
\caption{The time cost and perplexity of LoRA and EBFT on the LlamaV2-7B at sparsity of 20\%.}
\label{tab:time-cost}
\vspace{-0.7em}
\end{table} 
\begin{table*}[t]
\footnotesize
\begin{center}
  \begin{tabular}{ccccccccccc}
    \toprule
    \textbf{Model}&\textbf{Param.}&\textbf{Method}&\textbf{ARC-E}&\textbf{ARC-C}&\textbf{PIQA}&\textbf{WinoGrande}&\textbf{StoryCloze}&\textbf{Boolq}&\textbf{Mean}&\textbf{wiki.ppl}\\
    \midrule
    \multirow{5}{*}{\textbf{Lla.1}}&5.5B&LoRA&64.31&37.46&\textbf{76.66}&64.64&\textbf{77.28}&\textbf{71.47}&65.30&15.46\\
    \rowcolor{gray!=10}&5.5B&\textbf{Ours}&\textbf{72.52} &\textbf{38.65} &75.46 &\textbf{66.46} &75.63 &71.19 &\textbf{66.65}&\textbf{14.81}\\
    &5.0B&LoRA&60.48 &33.87 &\textbf{75.08}&61.80 &\textbf{76.16} & 63.82&61.87&16.67\\
    \rowcolor{gray!=10}&5.0B&\textbf{Ours}&\textbf{68.31} &\textbf{33.96} &72.85 &\textbf{63.85}&73.45 &\textbf{68.90}&\textbf{63.55}&\textbf{16.27}\\
    \midrule
    \multirow{5}{*}{\textbf{Lla.2}}&5.5B&LoRA&64.35&34.90&\textbf{75.84}&62.51&\textbf{75.47}&50.06&60.52&16.08\\
    \rowcolor{gray!=10}&5.5B&\textbf{Ours}&\textbf{68.81} &\textbf{35.24} &74.81&\textbf{63.93} &72.31 &\textbf{60.73} &\textbf{62.64} &\textbf{15.71}\\
    &5.0B&LoRA&61.32&32.08&\textbf{73.78}&61.96&\textbf{74.13}&55.05&59.72&\textbf{17.63}\\
    \rowcolor{gray!=10}&5.0B&\textbf{Ours}&\textbf{65.74} &\textbf{32.76}&71.87&\textbf{64.40}&71.04& \textbf{59.39}&\textbf{60.87}&\textbf{17.63}\\
    \bottomrule
  \end{tabular}
\end{center}
\caption{The accuracy and perplexity of the fine-tuned LlamaV1-7B and LlamaV2-7B models on Wikitext2, as well as their performance on a series of zero-shot tasks. The pruned models used in our experiments have parameters set at 5.5B and 5B, respectively.}
\label{tab:lora}
\end{table*}

We further conducted detailed experiments to compare our method with LoRA. We varied the parameters of the pruned models, including LlamaV1-7B and LlamaV2-7B, and evaluated the perplexity and accuracy of the fine-tuned models on Wikitext2 as well as a series of zero-shot tasks. The experimental results are summarized in Tab.\ref{tab:lora}. Indeed, the comparison between EBFT and LoRA continues to demonstrate the advantages of EBFT. For example, after fine-tuning LlamaV1-5.5B, EBFT achieves a perplexity of 14.81, surpassing LoRA, which achieves a perplexity of 15.46 on Wikitext2. Similarly, for LlamaV2-5.5B, EBFT achieves a perplexity of 15.71, outperforming LoRA with a perplexity of 16.08. This trend carries over to the zero-shot tasks as well, where the fine-tuned models using EBFT exhibit better performance compared to LoRA. The mean accuracy of our approach is higher than that of LoRA, regardless of whether it is applied to LlamaV1 or LlamaV2. While it is true that LoRA may achieve better scores on certain tasks such as PIQA and StoryCloze, the overall results consistently support the conclusion that the pruned models fine-tuned using EBFT outperform those fine-tuned using LoRA. \textbf{When comparing EBFT to LoRA}, EBFT demonstrates \textbf{faster speed}, \textbf{lower cost}, and \textbf{superior performance}.
\subsection{Weight Tuning vs. Mask Tuning}

Some optimization methods for sparse models, such as \cite{zhang2023lottery,zhang2023dynamic}, solely update the positions of masks without adjusting weights. To explore the effectiveness of this strategy, we conducted experiments to compare two fine-tuning strategies: weight tuning and mask tuning.

For mask tuning, we employed Eq.\ref{ourOO} as the optimization objective, aiming to minimize the block-wise reconstruction error. The fine-tuning process of mask tuning is the same as that of EBFT, except that mask tuning only updates the positions of masks while keeping the weights unchanged. We recorded the experimental results in Tab.\ref{tab:mask}. Specifically, we conducted variations in the sparsity levels of LlamaV1-7B and LlamaV2-7B, and evaluated the perplexity of the fine-tuned sparse models on Wikitext2. The results consistently highlight the clear advantage of weight tuning over mask tuning, even though the mask tuning method used in this study outperforms the SOTA mask-tuning method DSnoT in Tab.\ref{tab:un_wiki}. However, mask tuning still falls short when compared to EBFT. Regardless of the sparsity level, weight tuning consistently outperforms mask tuning. These findings clearly indicate the limitations of mask-tuning methods.
\begin{table}[t]
\footnotesize
\centering
\begin{tabular}{cccccc}
\toprule
\multicolumn{6}{c}{LlamaV1-7B}\\
\midrule
\multicolumn{1}{c|}{Method} &50\%&60\%&70\%&80\%&90\%\\
\midrule
 \multicolumn{1}{c|}{w.Mask}&7.05 &9.15&25.90&456.0&5378  \\
 \rowcolor{gray!=25}\multicolumn{1}{c|}{\textbf{w.Weight}} &\textbf{6.81} &\textbf{8.59}&\textbf{16.88}&\textbf{118.4}&\textbf{2993} \\
 \midrule
 \multicolumn{6}{c}{LlamaV2-7B}\\
\midrule
\multicolumn{1}{c}{Method} &50\%&60\%&70\%&80\%&90\%\\
\midrule
 \multicolumn{1}{c|}{w.Mask}&6.29 &8.40&26.99&755.8&3793  \\
 \rowcolor{gray!=25}\multicolumn{1}{c|}{\textbf{w.Weight}} &\textbf{6.18} &\textbf{7.90}&\textbf{16.94}&\textbf{72.80}&\textbf{903.4} \\
 
\bottomrule
\end{tabular}
\caption{The Wikitext2 perplexity of mask-tuning and weight-tuning were evaluated on LlamaV1-7B and LlamaV2-7B at various sparsity levels with Wanda initialization.}
\label{tab:mask}
\end{table}

\section{Conclusion}
We propose EBFT, a unified fine-tuning framework for sparse Language Models that can be integrated with any pruning method. In EBFT, we define the block-wise reconstruction error and optimize it on a block-by-block basis through backpropagation algorithm, aiming to achieve a convergent and optimal solution. This approach proves to be effective and efficient, requiring only a small number of samples for calibration. Extensive experiments demonstrate that EBFT achieves state-of-the-art performance on various benchmark datasets. 
\section{Limitation}
Although the use of a small calibration dataset significantly reduces costs, the fine-tuning process of EBFT still incurs computation costs due to gradient calculations. In future work, we will continue to focus on fine-tuning with a limited number of samples and explore gradient-free methods to further mitigate these costs.
\bibliography{anthology,custom}

\end{document}